\documentclass[letterpaper, 10 pt, conference]{ieeeconf}  

\IEEEoverridecommandlockouts                              

\title{\LARGE \bf
Robustness of Object Detectors in Degrading Weather Conditions
}

\author{Muhammad Jehanzeb Mirza$^{1,2}$, Cornelius Buerkle$^{3}$, Julio Jarquin$^{3}$, Michael Opitz$^{1,2}$, Fabian Oboril$^{3}$, \\ Kay-Ulrich Scholl$^{3}$, Horst Bischof$^{1,2}$
\thanks{$^{1}$Christian Doppler Laboratory for Embedded Machine Learning.}
\thanks{$^{2}$Institute for Computer Graphics and Vision, Graz University of Technology, Austria.}
\thanks{$^{3}$Intel Labs, Karlsruhe, Germany.} 
\thanks{This work was done at Intel Labs and later supported by Christian Doppler Laboratory for Embedded Machine Learning, Institute for Computer Graphics and Vision, Technical University of Graz. \newline
        {\tt\small Correspondence: Muhammad Jehanzeb Mirza <muhammad.mirza@icg.tugraz.at>}}
}

\usepackage{pgfplots}
\usepackage{amsmath,tikz,pgfplots}
\usepackage{tabularx}
\usepackage{multirow}
\usepackage{boldline}
\usepackage{verbatim}
\usepackage{hyperref}
\usepackage{graphicx}
\usepackage{amssymb}
\usepackage{pifont}
\begin{document}
\maketitle
\thispagestyle{empty}
\pagestyle{empty}

\begin{abstract}

State-of-the-art object detection systems for autonomous driving achieve promising results in clear weather conditions. However, such autonomous safety critical systems also need to work in degrading weather conditions, such as rain, fog and snow. Unfortunately, most approaches evaluate only on the KITTI dataset, which consists only of clear weather scenes. In this paper we address this issue and perform one of the most detailed evaluation on single and dual modality architectures on data captured in real weather conditions. We analyse the performance degradation of these architectures in degrading weather conditions. We demonstrate that an object detection architecture performing good in clear weather might not be able to handle degrading weather conditions. We also perform ablation studies on the dual modality architectures and show their limitations. 

\end{abstract}

\section{INTRODUCTION}

 Object detection is one of the most important perception problems in the context of autonomous driving. Object detectors rely on different sensors to perceive their environment. The most common sensors used for autonomous driving are are LiDAR and camera. Both sensors complement each other. The camera provides appearance information about the scene, such as color, texture, etc. In contrast, the LiDAR records sparse depth information about the scene.  3D detectors either use camera \cite{c13,c14,c15,c28,c29} or LiDAR \cite{c9,c10,c11} separately, or use fusion approaches to combine the information and to perform 3D object detection \cite{c16,c17,c18}.
\pgfplotstableread[row sep=\\,col sep=&]{
    interval & Total & Weather \\
    United States     & 5891000  & 1235000  \\
    Germany     & 2685661 & 214852  \\
    }\mydata
    
\begin{figure}
    \centering
    \includegraphics[scale=.55]{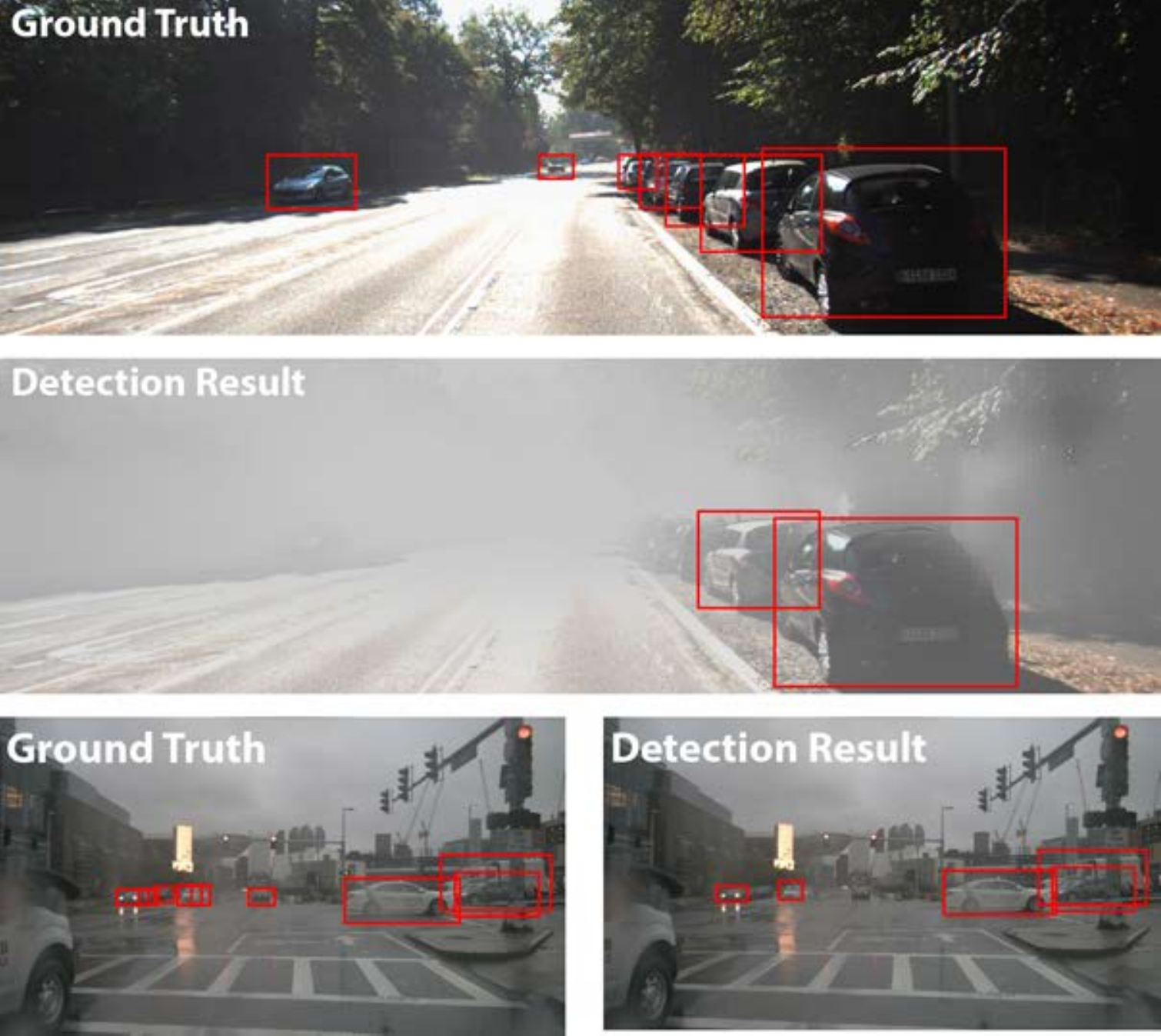}
    \caption{(Top two images)- Degradation in Augmented Fog (20m visibility) by employing \cite{c3} on KITTI Dataset \cite{c20}. (Bottom two images)- Degradation while testing with NuScenes \cite{c27}. Testing performed with YOLOv3 \cite{c7}.}
    \label{fig:qualitative_results}
\end{figure}

 In safety critical tasks such as autonomous driving, small mistakes can result in car crashes. Thus, it is necessary that the components, which are being used in an autonomous driving vehicle, are robust and fault tolerant. State-of-the-art multi-modal object detection architectures work reliably on current benchmark datasets, which are recorded in good weather conditions. However, for safety critical systems it is important that such systems work in adverse weather conditions in which sensors might fail. Specifically, 3D detectors should be able to cope with two important scenarios.

\begin{itemize}
    \item The performance should degrade gracefully with changing/degrading weather conditions. 
    \item The architectures should be robust against complete sensor failure scenarios caused e.g. by hardware failure or degrading weather. 
\end{itemize}

Unfortunately, capturing and annotating data in different weather conditions is expensive. Thus, there has been a lot of research regarding data augmentation e.g. \cite{c3,c4,c5,c6}  and evaluating object detection architectures on these augmented datasets. To the best of our knowledge, no current research provides a large scale study on robustness of present 2D and 3D object detectors in degrading weather conditions with data captured in real scenarios.  

We address this problem, and evaluate state-of-the-art single and multi-modal detection architectures in degrading weather conditions on real-world datasets in addition to augmented datasets. Specifically, we first evaluate single modality detectors (i.e. 2D and 3D detectors) in Section \ref{subsec:single_real} on autonomous driving datasets on a variety of different weather conditions. Further, we provide an extensive evaluation of a state-of-the-art multi-modal 3D detector \cite{c17} on these datasets in Section \ref{subsec:dual_real}. Finally, to show how these findings compare against evaluations performed purely on the synthetically created data from KITTI \cite{c20}, we also analyze the performance of these detection architectures on synthetically augmented data in Section \ref{subsec:single_aug}.

The remainder of this work is structured as follows. In Section \ref{sec:related_work} we give a brief overview of related work in the field. In Section \ref{sec:methodology} we discuss the datasets, detector architectures and evaluation details used in our experiments. Finally, in Section \ref{sec:eval} we present our extensive evaluation. 
\section{Related Work} \label{sec:related_work}

\subsection{2D Object Detection} 
Since the breakthrough with neural networks for image classification in 2012 by Krizhevsky et al. \cite{c19}, researchers adopted them for more complex computer vision tasks, such as 2D object detection for autonomous driving. 

We categorize 2D detection frameworks into two-step architectures, e.g. R-CNN based approaches \cite{c8,c21,c22,c46} and single-shot based detectors \cite{c7, c26, c26, c47, c48, c49, c50, c51, c52}. The two-step R-CNN based approaches were one of the first CNN based detection approaches. They typically achieve highly accurate detection results on popular benchmarks like KITTI \cite{c20} and Waymo  \cite{c44}. However, they first predict several object proposals, which they then refine in a second stage. Single-shot methods address this performance problem by directly predicting the bounding boxes.

\subsection{3D Object Detection}
In autonomous driving, it is important to estimate the distance to specific objects in the scene to safely navigate through the environment. Consequently, 3D detection is an increasingly important problem in this field. In this work, we categorize 3D detectors based on the modalities they use. Specifically, we distinguish between object detectors which use only a single modality, i.e. camera or LiDAR, or multiple modalities.
\subsection*{Single Modality 3D Object Detection}
A lot of research has been done by using only one of the modalities, i.e. either LiDAR or camera, and performing 3D (object) detection. Using only a single camera to predict 3D bounding boxes is an ill-posed problem because it is difficult to get the depth information from a single camera. However, several approaches address this problem, e.g. \cite{c13, c14,c15}. These methods typically use a multi-stage network. The first stage predicts several region proposals. The second stage then refines these proposals to 3D bounding boxes. Another line of work uses stereo cameras to estimate depth for 3D detection, e.g. \cite{c28, c29}.

Instead of using the camera as the modality, there is a line of work which uses only the 3D point cloud from LiDAR for 3D object detection. The main benefit of this approach is that these models can leverage the highly accurate 3D information of the LiDAR to predict 3D bounding boxes. Similar to camera based approaches, a popular line of work are two-stage approaches, e.g. \cite{c11,c24,c53,c54}. The first stage makes 3D object proposals. The second stage then refines these proposals to 3D bounding boxes. 
On the other hand there are detection networks that do not follow a multi-step approach. They are single shot detectors and detect bounding boxes in a single step e.g. \cite{c9,c10,c55,c56}.

\subsection*{Dual Modality 3D Object Detectors}

Another line of work uses both modalities (camera and LiDAR) for performing 3D object detection task. A number of popular fusion approaches have been proposed for such systems, which include, early, late or deep fusion \cite{c30}. In \cite{c16, c17} a multistage approach is used. The first stage predicts several regions and the second stage refines these proposals to predict the final 3D bounding box. In \cite{c18} researchers use a multi-task multi-sensor fusion architecture in which they show that sub tasks like depth completion and ground estimation are important tasks which can help increase the overall performance of the detection network.

In the context of autonomous driving these approaches are typically evaluated on the KITTI benchmark. Unfortunately, the KITTI benchmark only consists of scenes with good weather conditions. Consequently, it is impossible to assess the performance of such detectors on scenes with adverse weather conditions, e.g. rain, fog, snow and even during night conditions. In our work we address this problem and evaluate 3D and 2D object detectors on real-world datasets with a wide variety of different weather conditions. We provide details in Section \ref{subsec:datasets}.

\subsection{Effect on Sensors in Degrading Weather}
Understanding the effects of different weather conditions on the automotive sensors is an important task. There are studies which try to quantify the behavior of camera and LiDAR in different weather conditions such as rain, fog and snow, e.g. \cite{c31,c32,c34,c36,c37}.

Kshitiz et al. \cite{c35} give a detailed analysis of the effect of rain on the camera images. They further argue that the effect of rain strongly depends on the camera parameters. In successive works Kshitiz et al. \cite{c4,c5} propose a photo realistic rendering of rain streaks on the image by making use of mathematical models inspired by effects of rain in real environment. Garvelmann et al. \cite{c37} observe different snow processes by installing a large set of cameras. Spinneker et al. \cite{c36} propose a fog detection system for cameras which measures the blurriness of the image.

In contrast to camera, studying and quantifying the effect of different weather conditions on the LiDAR is more challenging because of the nature of the detection mechanism. However, studies have characterized the behavior of different weather conditions on LiDAR. In \cite{c31} the behavior of a variety of LiDARs has been tested by collecting the data in different snowy conditions. They conclude that echoes/false positives by the light beam hitting the snow flakes is not the same for all models of the LiDARs. They also conclude that the effect of snow is very limited for snowflakes falling at a range more than 10m. In \cite{c32} authors study the effect of rain on the LiDAR sensors. They conclude that range measurement can be affected by the light beams hitting the rain droplets. Kutila et al. \cite{c34} study the effect of fog and rain on LiDAR sensors and found that detection rate decreases by 50\% when weather becomes critical and visibility decreases.

These studies give a detailed overview of sensor behaviour in degrading weather conditions but do not address the effect on object detection. In contrast to these works, in this paper we want to quantitatively evaluate the degradation of 2D and 3D object detectors in adverse weather conditions.

\subsection{Augmenting Weather Conditions}\label{sec:data_aug}
One reason why there has not been enough research on testing the robustness of different perception algorithms in degrading weather conditions is because of the lack of data present in different weather conditions. Thus, a natural course of action is to employ realistic data augmentation pipelines. A lot of research has been done to augment camera images with rain, fog and snow as shown in \cite{c3,c4,c5,c6}. However, to our knowledge, realistic data augmentation pipelines to simulate different weather conditions on the LiDAR data are less known.

In our research we try to focus on evaluating object detection architectures on data captured in real weather conditions. However, to compare our findings with previous evaluation studies we also evaluate the object detection architectures on augmented data. To this end, we create augmented data with the data augmentation pipeline for camera images proposed by Halder et al. \cite{c3}. They render photo-realistic fog and rain on the camera images by making use of complex mathematical models which try to capture the effect of rain and fog in the real world.
\begin{figure}
    \centering
    \includegraphics[scale=.82]{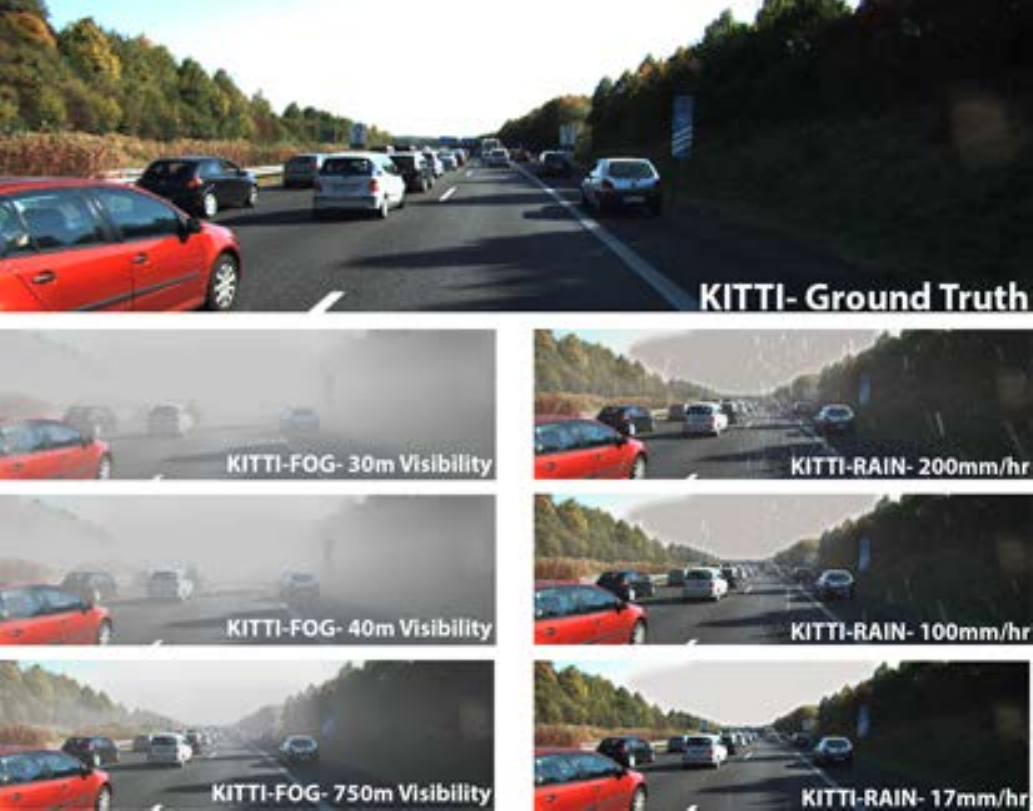}
    \caption{Data augmentation on the KITTI \cite{c20} by using \cite{c3}. \vspace{-2cm} }
    \label{fig:data_augmentation}
\end{figure}

\subsection{Evaluation Studies In Degrading Weather}
Although there has been a lot of work in characterising the behavior of different sensor modalities in a variety of weather conditions. Surprisingly, to our knowledge, there is little work found in literature where evaluation studies on object detection have been performed. 

In \cite{c39,c40} the performance of traffic sign detection in challenging weather has been analyzed. They primarily employ the CURE-TSR dataset \cite{c41} in which the data is augmented for simulating weather conditions or is captured from game engines. They show that as the weather conditions degrade, the detection performance drops. A more recent work \cite{c43} gives a detailed quantitative analysis of the performance of sensors in different weather conditions by using a controlled environment for their analysis. They also propose a deep fusion architecture for robust object detection. Further, Michaelis et al. \cite{c38} try to benchmark detection performance when the weather is degrading. This is close to our line of work in the autonomous driving domain. However, they only report 2D detection performance.

\section{Methodology} \label{sec:methodology}
We first describe the data we use for evaluating our object detection architectures. Then, we give an overview of the object detection architectures. Finally, we will describe the training and evaluation details.
\begin{table}
    \centering
    \begin{tabular}{ c|c|c|c|c } 
     \hlineB{4}
     Datasets& KITTI\cite{c20} &NuScenes\cite{c27} & Waymo\cite{c44} & A*3D\cite{c45} \\
     \hlineB{4}
     RGB& \ding{51} & \ding{51}&\ding{51} &\ding{51}\\ 
     LiDAR& \ding{51} &\ding{51} &\ding{51} &\ding{51}\\ \hline
     Day& \ding{51} &\ding{51} &\ding{51} &\ding{51}\\ 
     Night& \ding{55} &\ding{51} &\ding{51} &\ding{51}\\ 
     Rain& \ding{55} &\ding{51} &\ding{51} &\ding{55}\\ 
     Fog& \ding{55} &\ding{55} &\ding{55}&\ding{55}\\ 
     Snow& \ding{55} &\ding{55} &\ding{55}&\ding{55}\\
     \hlineB{4}
    \end{tabular}

    \caption{Datasets used in this paper and their characteristics.\vspace{-0.8cm}  }
    \label{tab:datasets}
\end{table}
\subsection{Datasets} \label{subsec:datasets}
Our main focus is on evaluating the object detection architectures on data which is collected in real scenarios. We provide an overview of the datasets, which we use in our evaluations in Table \ref{tab:datasets}. In Table \ref{tab:datasets} we see that different popular autonomous driving datasets consists only of scenes with limited variation in weather conditions. For example, the popular KITTI benchmark \cite{c20} consists of us with data captured during clear weather conditions. Further, none of the other popular multi-modal benchmark datasets consists of scenes with snow or fog. Snow and fog can create problems for the object detection architectures. For example, for LiDAR based object detectors they can yield false positive predictions \cite{c31,c34}. Further, camera based object detectors operating in these adverse weather conditions tend to miss detections \cite{c36,c38}. Consequently, it is not possible to assess the robustness of state-of-the-art detectors in adverse weather conditions on the KITTI benchmark. \cite{c20}.

We use four datasets as illustrated in Table \ref{tab:datasets}. They consist of 3 different data splits, which include: day, night and rain. We also evaluate the object detection architectures by augmenting fog and rain on the KITTI Dataset. For that purpose we use the open source implementation provided by Halder et al. \cite{c3}. This method augments photo-realistic fog and rain on images for different intensities by using the depth maps provided by KITTI. We create different augmented versions of KITTI by varying the fog visiblity and rain intensity parameters. We name these datasets KITTI-FOG and KITTI-RAIN in our paper. We illustrate an example of our augmented dataset with different visibility parameters in Fig. \ref{fig:data_augmentation}. We evaluate the degradation of object detectors on these datasets in Section \ref{subsec:single_aug}. 

\begin{table}
    \centering
    \begin{tabular}{ l|c|c|c } 
     \hlineB{4}
     Datasets& Total &Train Split & Validation Split\\
     \hlineB{4}
     KITTI\cite{c20}&7481 &3712 &3769 \\ 
     KITTI-RAIN&7481 &3712 &3769\\
     KITTI-FOG&7481 &3712 &3769\\ 
     NuScenes &38054 &18882 &19172\\ 
     NuScenes-DAY &25105 &12457 &12648\\ 
     NuScenes-NIGHT &3983 &1976 &2007\\ 
     NuScenes-RAIN &8966 &4449 &4517\\ 
     Waymo\cite{c44}&22500&15000&7500\\
     Waymo-DAY&22500&15000&7500\\
     Waymo-NIGHT&18148 &15000 &3148\\
     Waymo-RAIN&1193 &994 &199\\
     A*3D\cite{c45} &37665&18689&18976\\
     A*3D-DAY &27812&13800&14012\\
     A*3D-NIGHT &9853&4889&4964\\
     
     \hlineB{4}
    \end{tabular}

    \caption{Train and Validation samples used for evaluations.\vspace{-1cm} }
    \label{tab:train/eval_details}
\end{table}

\subsection{Object Detection Architectures}
There has been a significant amount of research on 2D and 3D object detection in the last years. We have mentioned some popular architectures in Section \ref{sec:related_work}. In our evaluation, we use 3 popular object detectors:

\begin{itemize}
    \item YOLOv3 \cite{c7} for evaluating the robustness of 2D object detection.
    \item PointPillars \cite{c9} for evaluating single modality 3D detection.
    \item AVODS \cite{c17} for evaluating multi-modal 3D detection.
\end{itemize}

These detectors achieve excellent performance on the challenging KITTI vision benchmark \cite{c20} while also achieving real-time runtime performance.

\subsection{Training/Evaluation Details}
We keep the training and evaluation details consistent for all the datasets. We follow previous work \cite{c16} and typically use a 50/50 split for training and validation for all datasets. In case of Waymo \cite{c44}, a dedicated validation and training split is provided, thus we use that. We then perform two types of evaluations: 
\begin{itemize}
    \item We evaluate all models on the full validation split and clear weather to provide a baseline performance of the detectors.
    \item We evaluate the trained detectors on specific scenarios (i.e. night, fog, rain) to provide detailed insights on the behaviour of our detectors in such adverse conditions.
\end{itemize}

We typically train all our models for 100 epochs on a single NVIDIA 2080Ti GPU with a constant batch size of 16.  However, when we train with \cite{c17}, we have to use a batch size of 1 due to memory constraints. We report the number of samples and the dataset splits in Table \ref{tab:train/eval_details}. 

\begin{table}
    \centering
    \begin{tabular}{ l|l|l|l|l } 
     \hlineB{4}
     Train Data&Validation Data&AP\_{car} &AP\_{ped} &mAP\_{@0.5} \\
     \hlineB{4}
     KITTI\cite{c20} &KITTI  &87.7 &67.3 &77.5\\ 
     KITTI &KITTI-RAIN&79.9 &58.1 &69\\
     KITTI &KITTI-FOG &45.2 &23.1 &34.2\\ \hline
     NuScenes\cite{c27} &NuScenes &73.2 &54.2 &63.7\\ 
     NuScenes &NuScenes-DAY &75.7 &57.8 &66.8\\ 
     NuScenes &NuScenes-NIGHT &50.2 &33.3 &41.7\\ 
     NuScenes &NuScenes-RAIN &69.1 &51.2 &60.2\\ \hline
     Waymo\cite{c44}&Waymo &40.9 &27.3 &34.1\\
     Waymo&Waymo-DAY &42.1 &28.9 &35.5\\
     Waymo &Waymo-NIGHT&26.9 &17 &21.9\\
     Waymo&Waymo-RAIN&32.5 &\_ &32.5\\ \hline
     A*3D\cite{c45}&A*3D&80.1 &50.6&65.4\\
     A*3D&A*3D-DAY&82.3 &55.5&68.9 \\
     A*3D&A*3D-NIGHT&53.4 &19.5&36.5\\
     
     \hlineB{4}
    \end{tabular}

    \caption{Single modality evaluation results by using Yolov3 \cite{c7}. \vspace{-0.4cm}}
    \label{tab:eval_yolo}
\end{table}

\begin{table}
    \centering
    \begin{tabular}{ l|l|l|l|l } 
     \hlineB{4}
    Train Data&Validation Data&AP\_{car} &AP\_{ped} &mAP\_{@0.5} \\
     \hlineB{4}
     KITTI\cite{c20} &KITTI  &84.7 & 48.2 &66.5\\\hline 
     NuScenes\cite{c27} &NuScenes &65.3 &47.1 &56.2\\ 
     NuScenes &NuScenes-DAY &65.9 &44.9 &55.4\\ 
     NuScenes &NuScenes-NIGHT &68.7 &47 &57.9\\ 
     NuScenes &NuScenes-RAIN &60.9 &45.3 &53.1\\ \hline
     Waymo\cite{c44}&Waymo &58.2 &47.1 &52.7\\
     Waymo&Waymo-DAY &59.7 &47.8 &53.8\\
     Waymo &Waymo-NIGHT&62.1 &47 &54.6\\
     Waymo&Waymo-RAIN&\_ &\_ &\_\\ \hline
     A*3D\cite{c45}&A*3D&73.1 &48.2 &60.7\\
     A*3D&A*3D-DAY&72 &48.8 &60.4\\
     A*3D&A*3D-NIGHT&73.1 &50.1 &61.6\\
     
     \hlineB{4}
    \end{tabular}

    \caption{Single modality evaluation results by using PointPillars \cite{c9}. \vspace{-1.2cm}}
    \label{tab:eval_pointpillars}
\end{table}
\section{Evaluations} \label{sec:eval}
In this section we provide a detailed evaluation of all our detectors in different weather conditions. 

We first evaluate our detectors on single modality real data. Then, we evaluate our detectors on augmented data. Finally, we evaluate a dual modality detector, i.e. AVODS \cite{c17}. 
\subsection{Single Modality Evaluations on Real Data} \label{subsec:single_real}
For our single modality experiments we use a state of the art 2D detector YOLOv3 \cite{c7} and a 3D detector PointPillars \cite{c9}. We show our results in Table \ref{tab:eval_yolo} and Table \ref{tab:eval_pointpillars}. Interestingly, there are three important effects we observe. 
\begin{itemize}
    \item The YOLOv3 (2D) detection accuracy degrades on night scenes compared to day scenes. Specifically mAP drops from 68.9 to 36.5 on the A*3D dataset, from 35.5 to 21.9 on the Waymo dataset and from 66.8 to 41.7 on the NuScenes dataset.
    \item Our LiDAR based 3D detector, i.e. PointPillars, achieves about 3\% improvement in mAP on night scenes. One possible explanation for this effect is, that LiDAR sensors tend to have a lower SNR during the day \cite{c57}.
    \item Both, 2D and 3D detectors decrease in about 6\% mAP in adverse weather conditions such as rain. We also illustrate this qualitatively in Fig. \ref{fig:qualitative_results}.  
\end{itemize}
\begin{table}
    \centering
    \begin{tabular}{ l|l|l|l|l } 
     \hlineB{4}
     Train Data&Validation Data&AP\_{car} &AP\_{ped} &mAP\_{@0.5} \\
     \hlineB{4}
     KITTI\cite{c20} &KITTI  &90.1 &43.6 &66.7\\\hline 
     NuScenes\cite{c27} &NuScenes &68.3 &40.1 &54.1\\ 
     NuScenes &NuScenes-DAY &67.2 &37.2 &52.2\\ 
     NuScenes &NuScenes-NIGHT &69.8 &40.3 &55.1\\ 
     NuScenes &NuScenes-RAIN &63.1 &35.1 &49.1\\ \hline
     Waymo\cite{c44}&Waymo &59.1 &42.3 &50.7\\
     Waymo&Waymo-DAY &58.9 &42 &50.5\\
     Waymo &Waymo-NIGHT&61.7 &43.1 &52.4\\
     Waymo&Waymo-RAIN&\_ &\_ &\_\\ \hline
     A*3D\cite{c45}&A*3D&75.2 &33.1 &54.2\\
     A*3D&A*3D-DAY&71.5 &38.5 &55\\
     A*3D&A*3D-NIGHT&77.9 &39.3 &58.6\\
     \hlineB{4}
    \end{tabular}
    
    \caption{Dual modality evaluation results by using AVOD \cite{c17}.\vspace{-0.4cm} }
    \label{tab:eval_avods}
\end{table}

\begin{table}
    \centering
    \begin{tabular}{ l|l|l|l|l } 
     \hlineB{4}
     Modality&Validation Data&AP\_{car} &AP\_{ped} &mAP\_{@0.5} \\
     \hlineB{4}
     LiDAR&A*3D-DAY&71.1 &42.2 &56.7\\
     Camera&A*3D-DAY&0 &0 &0\\
     LiDAR&A*3D-NIGHT&78.2 &41.9 &60.1\\
     Camera&A*3D-NIGHT&0 &0 &0\\\hlineB{3}
     
     \hlineB{4}
    \end{tabular}
    
    \caption{Results for Ablation Studies on AVOD \cite{c17}.\vspace{-0.9cm} }
    \label{tab:eval_ablation}
\end{table}
\subsection{Single Modality Evaluations on Augmented Data} \label{subsec:single_aug}

\begin{figure}
    \centering
    \begin{tikzpicture}
    \begin{axis}[
        width=7cm,
        height = 5cm,
        xlabel={Rain Intensity (mm/hr)},
        ylabel={Mean Average Precision (\%)},
        xmin=0, xmax=200,
        ymin=0, ymax=100,
        xtick={0,40,80,120,160,200},
        ytick={0,20,40,60,80,100},
        legend pos=south east,
        ymajorgrids=true,
        grid style=dashed,
    ]
    
    \addlegendentry{KITTI-RAIN}
    \addplot[
        color=blue,
        mark=triangle,
        ]
        coordinates {
        (17,75.30)(50,73.10)(75,72.25)(100,69.36)(200,66.74)
        };
    \addplot[mark=none,red,thick] coordinates {(0, 76.80) (100, 76.80) (200, 76.80) (800, 76.80)};
    \addlegendentry{KITTI}
        
    \end{axis}
    \end{tikzpicture}
    \caption{Evaluation results while testing on Augmented Rain. \vspace{-5cm}}
    \label{fig:KITTI_rain_results}
\end{figure}
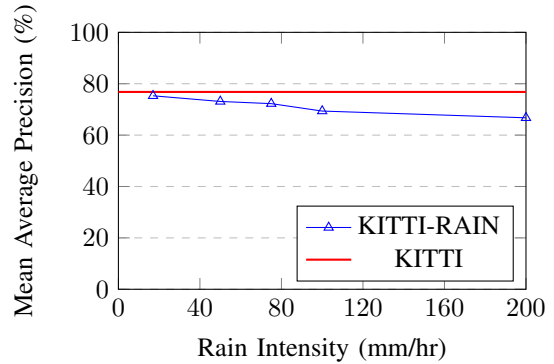

We also evaluate our 2D object detector, i.e. YOLOv3 \cite{c7}, by synthetically augmenting  the KITTI  dataset \cite{c20} with different rain and fog intensities. Consequently, we can test how well an object detector trained with the standard KITTI dataset can cope with different weather conditions. We summarize our findings in Fig. \ref{fig:KITTI_fog_results} and Fig. \ref{fig:KITTI_rain_results}. We see that as fog visibility decreases, the detector mAP significantly decreases. Further, by increasing the rain intensity, the mAP of our detector decreases as well. 

Consequently, strong performance on the original KITTI benchmark does not translate to a robust performance in adverse weather conditions.

\subsection{Dual Modality Evaluations}\label{subsec:dual_real}
We also evaluate the state-of-the-art dual modality architecture AVOD \cite{c17} on different datasets. We report the accuracy in Table \ref{tab:eval_avods}.  
\begin{itemize}
    \item Interestingly, AVOD does not achieve better results compared to PointPillars \cite{c9}, which uses only the LiDAR modality as input. In fact some results obtained are even worse than using only the LiDAR only method. E.g. When we evaluated on the Waymo \cite{c44} data splits, the results achieved by PointPillars are better than the results achieved by using the dual modality architecture.
    
    \item Further, we form a hypothesis that the dual modality architecture used for evaluation in this paper relies more on LiDAR data. This is supported by comparing the results obtained by the single modality and dual modality architectures in Table \ref{tab:eval_pointpillars} and Table \ref{tab:eval_avods}.  
\end{itemize}

\subsection{Ablation Studies}\label{sub:ablation}
In this section we perform ablation studies on a multi modal architecture, i.e. AVOD \cite{c17}. We test the robustness of this architecture in the event of complete sensor failure. Therefore, we try to omit the LiDAR or camera input completely during our experiments. Such an event could occur due to e.g. hardware failure, degrading weather, etc. We summarize the results in Table \ref{tab:eval_ablation}.
\begin{figure}
    \centering
    \begin{tikzpicture}
    \begin{axis}[
        width=7cm,
        height = 5cm,
        xlabel={Maximum Visibility (m)},
        ylabel={Mean Average Precision (\%)},
        xmin=0, xmax=800,
        ymin=0, ymax=100,
        xtick={0,100,200,300,400, 500, 600,700,800},
        ytick={0,20,40,60,80,100},
        legend pos=south east,
        ymajorgrids=true,
        grid style=dashed,
    ]
    
    \addlegendentry{KITTI-FOG}
    \addplot[
        color=blue,
        mark=triangle,
        ]
        coordinates {
        (30,33.2)(40,37.4)(75,51)(750,75.1)
        };
    \addplot[mark=none,red,thick] coordinates {(0, 76.80) (100, 76.80) (200, 76.8) (800, 76.80)};
    \addlegendentry{KITTI}
        
    \end{axis}
    \end{tikzpicture}
    \caption{Evaluation results while testing on Augmented Fog.}
    \label{fig:KITTI_fog_results}
\end{figure}
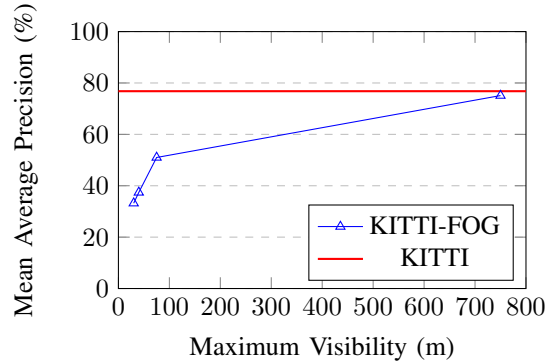

In our evaluations we see that dual modality detectors fail when there is no LiDAR data available, but tend to continue working when there is no camera data available. One reason for this might be that the region proposal network of the AVOD detector relies on the LiDAR modality to create 3D proposals. The data from the camera is only used for refinement of the proposals. Consequently, such detectors fail with missing LiDAR data.  

\section{Discussion and Conclusion \vspace{-.1cm}}
Our detailed evaluation in Table III, IV and V summarizes the state-of-the-art in object detection for autonomous driving in adverse weather conditions. From our results we conclude that the detectors which perform good on the KITTI benchmark \cite{c20}, can fail in such adverse weather conditions. The main reason for this is that this benchmark does not cover such extreme weather scenarios. Consequently, it is necessary to improve benchmarking datasets to cover such cases. 

Another takeaway is that the design of multi-modal detectors should take sensor degradation into account. In order to avoid that the entire detector becomes overly reliant on a specific input modality. As our results with AVOD \cite{c17} show, such a biased-design can lead to poor results under an adverse scenario where the LiDAR modality is failing. For safety critical systems it is necessary that such detectors are fault tolerant. As part of the future work we will be moving towards designing realistic data augmentation pipelines to augment different weather conditions on the LiDAR point clouds. Further, we will design a fault tolerant object detection architecture which can perform in the case of a complete sensor failure, whether it is camera sensor failure or a LiDAR sensor failure.

\section*{Acknowledgment}
The financial support by the Austrian Federal Ministry for Digital and Economic Affairs, the National Foundation for Research, Technology and Development and the Christian Doppler Research Association is gratefully acknowledged.

Christian Doppler Laboratory for Embedded Machine Learning, Institute for Computer Graphics and Vision, Graz University of Technology, Austria.

The computational results presented have been achieved [in part] using the Vienna Scientific Cluster (VSC).






\end{document}